\documentclass[letterpaper, 10 pt, conference]{ieeeconf}  

\IEEEoverridecommandlockouts                              

\usepackage[T1]{fontenc}
\usepackage[utf8]{inputenc}
\usepackage{authblk}
\usepackage{graphicx}
\usepackage{array}
\usepackage{float}
\usepackage{float}
\usepackage{booktabs}
\usepackage{bbding}
\usepackage{etoolbox}
\usepackage{color}
\usepackage{soul}
\usepackage{algorithm}
\usepackage{algorithmic}
\usepackage{cite}
\usepackage{balance}
\usepackage{amsmath, amssymb}
\usepackage{makecell}
\usepackage{multirow}
\usepackage{hyperref}

\begin{document}

\title{\LARGE \bf
OSSAR: Towards Open-Set Surgical Activity Recognition in Robot-assisted Surgery
}

\author{Long Bai$^{1,\dagger}$, Guankun Wang$^{1,\dagger}$, Jie Wang$^{2}$, Xiaoxiao Yang$^3$, Huxin Gao$^{1}$, Xin Liang$^{4}$, An Wang$^{1}$, \\ Mobarakol Islam$^{5}$, and Hongliang Ren$^{1,6,*}$, \textit{Senior Member, IEEE}
\thanks{*This work was supported by Hong Kong RGC CRF C4026-21GF, GRF 14203323, GRF 14216022, GRF 14211420, NSFC/RGC Joint Research Scheme N\_CUHK420/22, Shenzhen-Hong Kong-Macau Technology Research Programme (Type C) STIC Grant 202108233000303. (Corresponding author: H. Ren, hlren@ieee.org)}
\thanks{$^{\dagger}$ Co-first authors.}%
\thanks{$^1$ L. Bai, G. Wang, H. Gao, A. Wang, and H. Ren are with the Dept. of Electronic Engineering, The Chinese University of Hong Kong (CUHK), Hong Kong, China. (E-mail: b.long@ieee.org, gkwang@link.cuhk.edu.hk)}
\thanks{$^2$ J. Wang is with Beijing Institute of Technology, Beijing, China.}
\thanks{$^3$ X. Yang is with the Qilu Hospital of Shandong University, Jinan, China.}
\thanks{$^4$ X. Liang is with Tongji University, Shanghai, China.}
\thanks{$^5$ M. Islam is with the Wellcome/EPSRC Centre for Interventional and Surgical Sciences (WEISS), University College London, London, UK.}
\thanks{$^6$ H. Ren is also with the Dept. of Biomedical Engineering, National University of Singapore, Singapore.}
}


\maketitle
\thispagestyle{empty}
\pagestyle{empty}

\begin{abstract}

In the realm of automated robotic surgery and computer-assisted interventions, understanding robotic surgical activities stands paramount. Existing algorithms dedicated to surgical activity recognition predominantly cater to pre-defined closed-set paradigms, ignoring the challenges of real-world open-set scenarios. Such algorithms often falter in the presence of test samples originating from classes unseen during training phases. To tackle this problem, we introduce an innovative \textit{Open-Set Surgical Activity Recognition} (OSSAR) framework. Our solution leverages the hyperspherical reciprocal point strategy to enhance the distinction between known and unknown classes in the feature space. Additionally, we address the issue of over-confidence in the closed set by refining model calibration, avoiding misclassification of unknown classes as known ones. To support our assertions, we establish an open-set surgical activity benchmark utilizing the public JIGSAWS dataset. Besides, we also collect a novel dataset on endoscopic submucosal dissection for surgical activity tasks. Extensive comparisons and ablation experiments on these datasets demonstrate the significant outperformance of our method over existing state-of-the-art approaches. Our proposed solution can effectively address the challenges of real-world surgical scenarios. Our code is publicly accessible at \href{https://github.com/longbai1006/OSSAR}{github.com/longbai1006/OSSAR}.

\end{abstract}

\section{INTRODUCTION}
\label{sec:intro}
Activity recognition in robot-assisted surgery has emerged as a focal research avenue, primarily attributed to its promising implications for intelligent computer-assisted intervention and automatic robotic surgery~\cite{wagner2023comparative}. It entails the development and application of algorithms and techniques aimed at comprehending and categorizing the various stages or steps involved in surgical procedures~\cite{maier2022surgical}. 
By performing a series of assistive tasks (e.g., monitoring surgical procedures~\cite{maier2022surgical}, facilitation in decision-making and communication~\cite{forestier2015automatic,demir2023deep}, providing warning information~\cite{maier2017surgical}, and offering surgical educational resources~\cite{ortenzi2023novel}), surgical activity recognition (SAR) paves the way for the overall efficiency, safety, and precision of robot-assisted surgery, thereby promoting the development of intelligent surgical robotic systems~\cite{maier2017surgical,bai2023cat}. However, vision-based SAR also confronts great challenges due to the complexity of surgical scenes, variations among different surgeon and patient cohorts, and the appearance similarity between classes~\cite{jin2022trans,wang2023rethinking,bai2023llcaps}.

\begin{figure}[t]
    \centering
    \includegraphics[width=0.7\linewidth, trim=0 250 755 0]{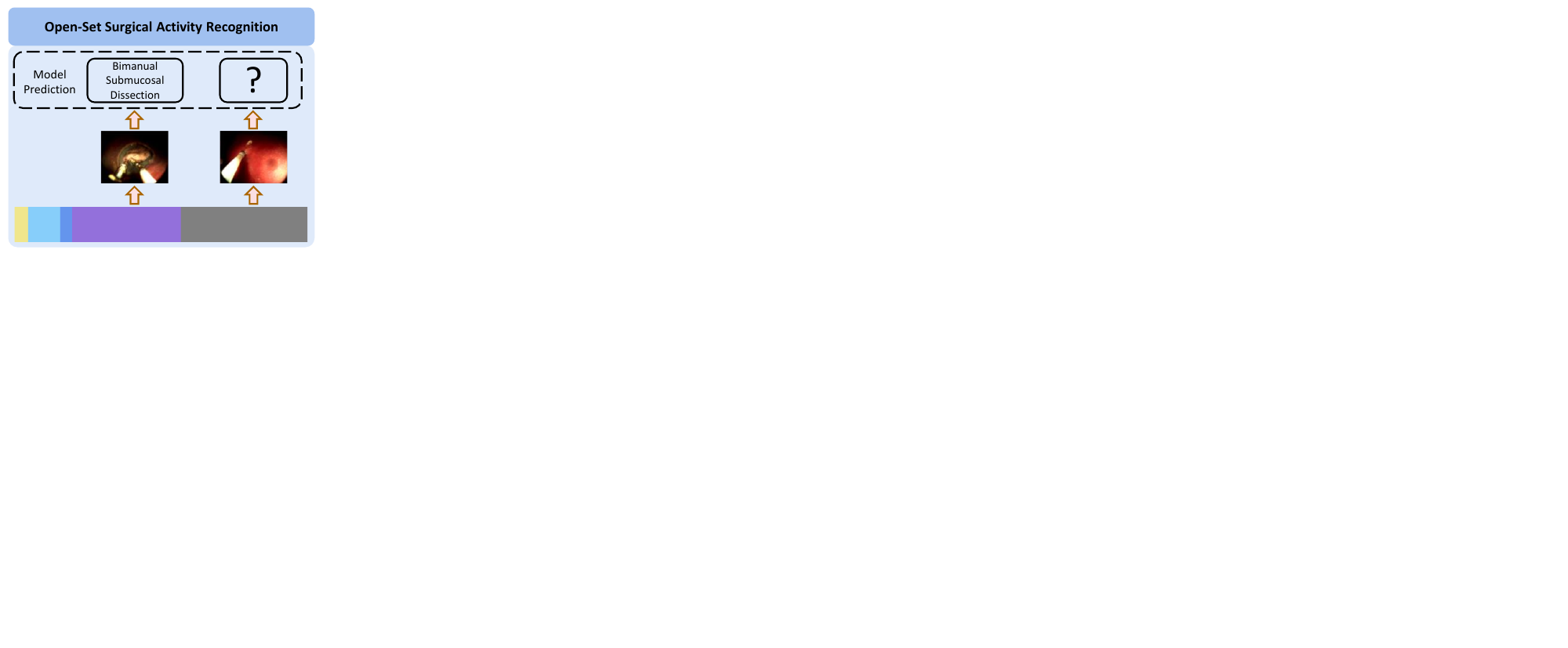}
    \caption{Illustration of open-set surgical activity recognition. The gray band depicts the unknown classes, whereas the others represent the known classes. The model is able to predict the pre-defined classes, but if the class is not defined during the training stage, the model cannot discriminate it during testing.}
    \label{fig:intro}
\end{figure}

Nevertheless, the practicality of real-world situations often entails a multitude of scenarios that remain unencountered during the model's training phase, posing a substantial challenge to the generalizability of contemporary deep learning methodologies~\cite{bendale2016towards,che2022learning,jin2023unsupervised,che2023towards,wang2023domain}. Consequently, enhancing the resilience of surgical activity analysis in the face of previously unseen data represents an imperative yet intricate research problem.
As shown in Fig.~\ref{fig:intro}, the intricacy of the surgical activity identification problem significantly increases due to the existence of unseen scenarios or categories. Specifically, in the domain of surgical activity analysis, it is observed that categories are not always a priori-defined. The proliferation of robot-assisted surgical procedures has led to the inception of novel categories. These emergent categories can potentially intersect with and obscure traditional classifications, thereby complicating precise classification. Hence, there is an urgent need to develop an open-set recognition (OSR) solution tailored to surgical activities to address the challenges arising from the open-world scenario.

OSR tasks have been discussed extensively in the computer vision community. To enhance the model's resistance to unknown classes, an advanced technique in prior studies is adversarial reciprocal point learning (ARPL)~\cite{chen2020learning, chen2021adversarial}.
These techniques commonly infuse extraneous information into learners restricted to known class data, aiming to derive compact and discriminative representations. Notwithstanding their efficacy, there exists one predominant shortcoming. 
In scenarios where the divergence between known and unknown data remains marginal, these methods demonstrate compromised discriminative capabilities, resulting in potential misclassifications~\cite{deng2019arcface,peng2022angular}.

To tackle the issues at hand, we propose the Open-Set Surgical Activity Recognition (OSSAR) framework, consisting of two key components: the HyperSpherical Reciprocal Points (HSRP) strategy and the Closed-set Over-confidence Calibration (COC) module. The HSRP strategy is tailored for handling unknown classes, leveraging the benefits of hyperspherical metric space. By exploiting the invariant characteristics of angular distance, we effectively identify subtle visual representations with minuscule inter-class differences and push the unknown classes away. On the other hand, the COC loss is applied to the closed-set, efficiently penalizing and mitigating the model's excessive confidence in known classes, thus mitigating the misclassification of similar unknown samples as known ones. Our framework demonstrates its effectiveness with superior performance on two benchmark datasets. Specifically, our contributions can be summarized as follows:
\begin{itemize}
    \item We present a new open-set surgical activity recognition (OSSAR) framework, which is tailored for open-set challenges in the presence of real-world surgical activity recognition.
    \item We propose the hyperspherical reciprocal point strategy to learn discriminative features on a hypersphere space. Besides, we employ a over-confidence calibration mechanism that can mitigate misclassification issues for similar samples by incorporating conditional constraints.
    \item We construct the OSSAR benchmark setup on the public JIGSAWS dataset, which can facilitate research on real-world surgical activities. We also establish another comprehensive robot-assisted surgery activity dataset on endoscopic submucosal dissection (ESD). We demonstrate the effectiveness of our method through extensive comparison and ablation experiments on the two datasets, showcasing significant improvements in model generalization compared to SOTA methods.
\end{itemize}

\section{Related Work}
\label{sec:relate}
\subsection{Learning on Surgical Activity Recognition}
\label{sec:related_sar}

The domain of robot-assisted surgery has witnessed a profound evolution with the advent of computational methods and intelligent systems. 
In the early stages, statistical models such as Hidden Markov Models (HMM) were primarily used to classify surgical steps. With deep learning solutions emerging, image classification strategies with temporal convolutional networks (TCN)~\cite{sanchez2022data}, recurrent convolutional neural networks (RCNN)~\cite{jin2020multi} or long short-term memory (LSTM)~\cite{rivoir2020rethinking} enable frame-by-frame prediction of surgical activities and stages. Recently, Transformer has gained popularity in SAR tasks due to its exceptional ability to capture long-range information~\cite{jin2022trans}. 
However, these methods struggle to handle the open-world challenge of undefined categories in surgical data, as they heavily rely on pre-defined closed-set classifications. Besides, in the surgical domain, Jan Sellner \textit{et al.}~\cite{sellner2023semantic} have discussed the impact of out-of-distribution data caused by geometric domain shift, while Maya Zohar \textit{et al.}~\cite{pmlr-v121-zohar20a} proposed to detect out-of-body data to address privacy concerns during surgical procedures. Nevertheless, the open-set scenario in SAR tasks that involves anomalous or undefined categories remains unresolved and requires further scholarly exploration.

\begin{figure*}[t]
    \centering
    \includegraphics[width=0.92\linewidth, trim=0 20 80 0]{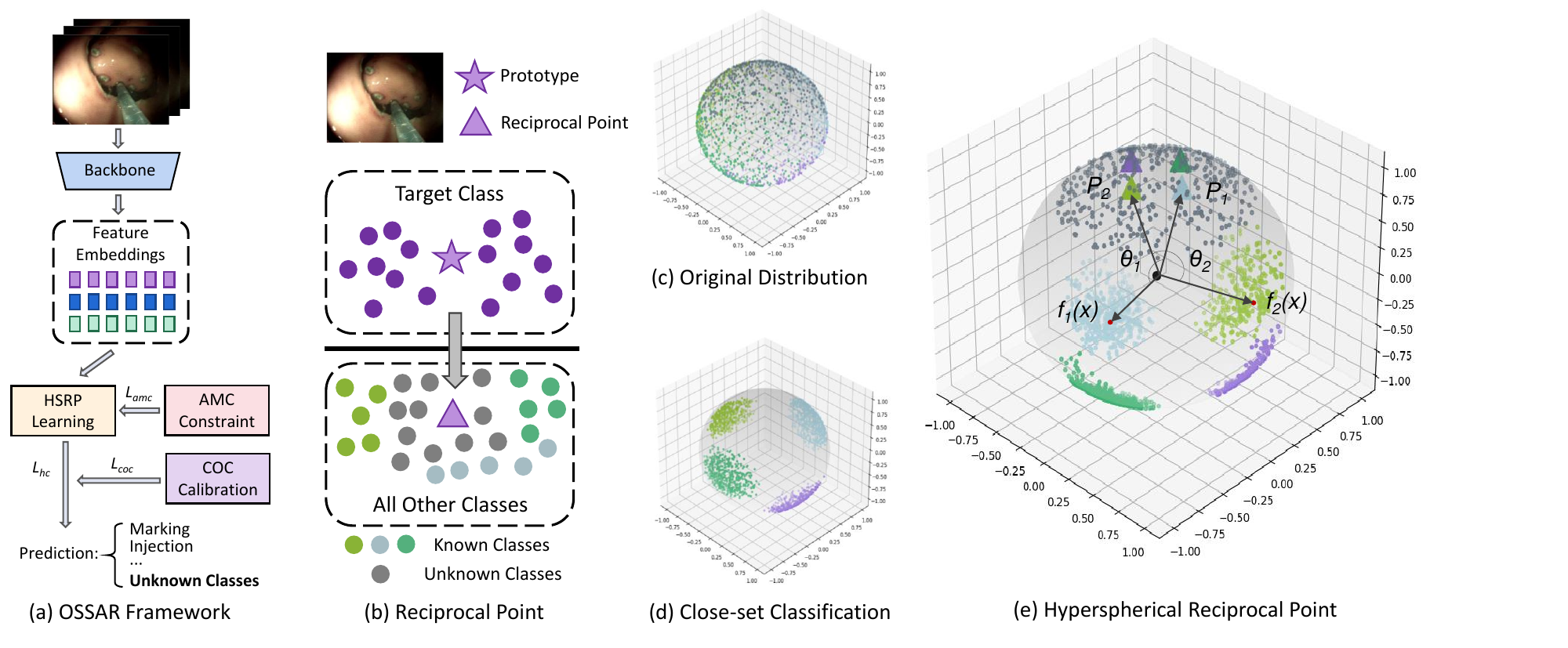}
    \caption{Overview of our OSSAR framework. (a) The general procedure of our OSSAR framework is demonstrated, with the HSRP classification loss $\mathcal{L}_{hc}$, the adversarial margin constraint loss $\mathcal{L}_{amc}$, and the closed-set over-confidence calibration loss $\mathcal{L}_{coc}$. (b) The principle of reciprocal points is demonstrated, in which the opposite points of target classes are used to represent the feature space of all known and unknown classes except for the target class. (c) The disordered state of the feature space, when it is unclassified in the hyper-spherical feature space, is presented. (d) The ordinary closed-set classification scenario is shown, where different classes are clustered together. (e) Our hyperspherical reciprocal point solution is presented, which pushes the unknown classes closer to the reciprocal points, resulting in better performance of unknown class detection.}
    \label{fig:main}
\end{figure*}

\subsection{Advancements in Open-Set Recognition}
\label{sec:related_osr}
OSR poses a profound challenge in the computer vision discipline, tackling the intricacies of scenarios wherein test samples might hail from unobserved classes during the training phase. Scheirer \textit{et al.}~\cite{scheirer2012toward} formulated the OSR problem, introducing a framework that hinged upon the support vector machine (SVM) model. Bendale \textit{et al.}'s seminal work~\cite{bendale2016towards} marked the adaptation of OSR to deep learning frameworks, employing the Extreme Value Theory (EVT) to extend the capabilities of the K-class softmax classifier. G-Openmax~\cite{ge2017generative, moon2022difficulty} incorporate a generative model to augment the OSR capabilities of deep neural networks, training them with synthetically generated unknown samples. Through a reconstructive lens, CROSR~\cite{yoshihashi2019classification} harnesses the power of self-supervised learning to reconstruct known class data representations, facilitating the discernment of unknown classes. Concurrently, metric-based and prototype learning methodologies~\cite{chen2020learning, chen2021adversarial, yang2020convolutional, zhang2021prototypical}, epitomized by RPL~\cite{chen2020learning}, endeavor to discern unknown entities by generating significant distances relative to known class data prototypes.
Vaze \textit{et al.}~\cite{vaze2021open} identify an explicit association between a model's performance under closed-set conditions and its proficiency in OSR scenarios.
Furthermore, DEAR~\cite{bao2021evidential} employs deep evidential learning to formulate open-set action recognition as a problem of uncertainty quantification.
While these methodologies have exhibited excellent performance in addressing OSR challenges, there remains a potential oversight pertaining to the issue of over-confidence, which can precipitate the misclassification of akin samples. In this manuscript, we present an advanced confidence-constraint mechanism explicitly designed to mitigate this challenge.

\section{METHODOLOGY}
\label{sec:method}

\subsection{Overview}
\label{sec:overview}
In the open-set activity recognition, given the training samples $(x_i,y_i) \in \mathcal{D}_I$, the training set $\mathcal{D}_I$ contains $K$ known classes (also called as inliers). The test set $\mathcal{D}_T$ includes $K$ known classes and $U$ unknown classes (also called outliers). $F(x)$ denotes the embedding learning function. Following previous works on surgical vision tasks~\cite{bai2023revisiting,jin2022exploring,bai2023surgical,anastasiou2023keep}, we adopt the ResNet-18~\cite{he2016deep} as the backbone to conduct feature learning and extract $F(x)$.
During the training stage, our proposed HSRP strategy processes $F(x)$ in the hyperspherical space and simulates the presence of unknown classes. The adversarial margin constraint (AMC) module imposes a constraint on HSRP, ensuring the objective is optimized within a bounded space. Subsequently, the classification loss is executed, and the COC module is applied to the classification loss to calibrate the over-confidence. The overview is presented in Fig.~\ref{fig:main}~(a).

\subsection{Preliminaries}
\label{sec:preli}

Adversarial reciprocal points learning (ARPL)~\cite{chen2021adversarial} proposes the reciprocal points $\mathcal{P}_C = \left\{p_i^c \mid i=1, \ldots, N\right\}$, an $N$-dimensional representation, as the latent representation for all classes in $\mathcal{D}_I \cup \mathcal{D}_T$ except for class $C$, as presented in Fig.~\ref{fig:main}~(b). ARPL seeks to amplify the distance between $\mathcal{P}_C$ and the learned embeddings $F(x)$ of corresponding known samples. The distance is calculated by $\mathrm{Dist}(F(x), \mathcal{P}_C)= \left\|F(x)-p_i^c\right\|_2^2 /N - F(x) \cdot \mathcal{P}_C $.
The reciprocal classification loss can be formulated as:
\begin{equation}
    \mathcal{L}_{c}(x;\mathcal{P})= -\mathrm{log}\frac{\mathrm{exp}\left[{\tau \mathrm{Dist}\left(F(x), \mathcal{P}_C\right)}\right]}{\sum_{i=1}^K \mathrm{exp}\left[{\tau \mathrm{Dist}\left(F(x), \mathcal{P}_i\right)}\right]}
\end{equation}
The hardness parameter $\tau$ is set to $1$ following~\cite{chen2021adversarial}.
Furthermore, to address the open space risk in OSR, ARPL incorporates adversarial margin constraints (AMC) between reciprocal points and known classes. The AMC loss is defined as:
\begin{equation}
    \mathcal{L}_{amc}\left(x ; \mathcal{P}, R\right)=\max \left(\|F(x), \mathcal{P}_C\|^2_2 \right)-R, 0)
\end{equation}
Finally, with $\alpha$ as the weight, the ARPL loss is defined as:
\begin{equation}
    \mathcal{L}_{ARPL}=\mathcal{L}_{c}(x ; \mathcal{P})+ \alpha \mathcal{L}_{amc}(x ; \mathcal{P}, \mathcal{R})
\end{equation}

In addition to its core functionalities, ARPL incorporates a generative mechanism to synthesize confusing samples, which further maximizes the divergence between known and unknown classes. However, based on our empirical study in Section~\ref{sec:results}, it is difficult to generate samples with distinguishable features for surgical data. Therefore, we elect to omit the confused samples from our framework.

\subsection{Hyperspherical Reciprocal Points}
\label{sec:hsrp}
Referring to Fig.~\ref{fig:main}, we propose to tackle the OSR problems through hypersphere learning. Fig.~\ref{fig:main}~(c) and (d) respectively reveal the distribution of the hyperspherical feature space for the unclassified and closed-set classification scenarios. Fig.~\ref{fig:main}~(e) demonstrates how we employ the hypersphere to handle reciprocal points. In surgical scenarios where different classes exhibit highly similar features, we leverage hyperspherical representation to enhance the cohesiveness of samples within a class and to augment the distinction between representations of different classes. The angular distance in the hypersphere is formulated as: 
\begin{equation}
\mathrm{Dist}'\left(F(x), P_C\right)=\cos \theta_{x,\mathcal{P}}=\frac{F(x)^T P_C}{\|F(x)\|\left\|P_C\right\|}
\end{equation}
Compared to computing the difference between Euclidean distance and dot product, the angular distance can directly measure the angle $\theta$ between vectors, which is a more concise, effective, and efficient method. Furthermore, the invariance of angular distance enables the model to handle similar feature representations effectively.
By exploring feature space distribution on the 3D hypersphere, we can achieve robust prediction and uncertainty estimation for known classes and their corresponding reciprocal points.
The HSRP classification loss shall be formulated as:
\begin{equation}
    \mathcal{L}_{hc}(x;\mathcal{P})= -\mathrm{log}\frac{\mathrm{exp}\left[{\tau \cdot \cos( \theta_{x,\mathcal{P}})}\right]}{\sum_{i=1}^K \mathrm{exp}\left[{\tau \cdot \cos( \theta_{x,\mathcal{P}})}\right]}
\end{equation}

Although HSRP exhibits the capability to expand the feature space and alleviate the challenges in optimizing the classification loss, the angular loss remains vulnerable to the issue of insensitivity to scaling.
In contrast, the Euclidean distance is effective in handling the scaling transformations of feature embeddings and reciprocal points, given its sensitivity to each vector component. Hence, we still utilize the Euclidean distance to impose margin constraints. Our empirical results, as depicted in Table~\ref{tab:distance}, showcase that the angular distance in $\mathcal{L}_{amc}$ does not present a better performance. Despite its closed-set accuracy aligning comparably with that of ARPL, $\mathcal{L}_{amc}$ with the angular distance evidences a pronounced decline in efficacy when identifying unknown classes. Finally, the total loss of our HSRP strategy can be rewritten to:
\begin{equation}
    \mathcal{L}_{HSRP}=\mathcal{L}_{hc}(x ; \mathcal{P})+ \alpha \mathcal{L}_{amc}(x ; \mathcal{P}, \mathcal{R})
\end{equation}

\subsection{Closed-set Over-confidence Calibration}
\label{sec:coc}
The correlation between closed-set accuracy and OSR performance has been demonstrated in existing literature~\cite{vaze2021open}. When a model achieves excellent accuracy on the closed-set, it is more likely to detect unknown classes effectively during inference. Nevertheless, during various stages of robotic surgery, the explicit features are highly similar, including similarities in the target organs, surgical instruments, and surroundings. Therefore, the model is more prone to misclassify unknown classes as known, leading to a decline in OSR performance.

In this case, we attempt to calibrate and penalize the over-confidence on the closed-set during the training period. As discussed in Section~\ref{sec:hsrp}, we replace the commonly used logits with our angular distance $\cos( \theta_{x,\mathcal{P}})$ and conduct the cross-entropy convergence based on it. The miscalibration model tends to implicitly amplify the distances $\lambda$ between the highest logit and the remaining ones, resulting in an overconfident prediction~\cite{szegedy2016rethinking,pereyra2017regularizing,liu2022devil}. Given an input image belonging to class $C \in [1, K]$, its angular distance can be denoted by $\cos( \theta_{x,\mathcal{P}})_C$. Thus, the $K$-dimensional $\lambda$ can be formulated by the distance between the class with the highest distance and the others:
\begin{equation}
    \lambda_C(x;\mathcal{P}) = \max _j\left[\cos( \theta_{x,\mathcal{P}})_j\right]-\cos( \theta_{x,\mathcal{P}})_C
\end{equation}

When pushing $\lambda_C$ towards $0$, the output probabilities for each class will be close to $1/K$, which is unreasonable and lacks useful information. Consequently, a positive constrained threshold $\theta$ can be set to constrain $\lambda_C \leq \theta$. The constrained threshold $\theta$ can regulate and penalize the backpropagation of gradients. 
Backpropagation of the gradient is allowed only when the distance exceeds our prespecified threshold $\theta$.
We can map it with a ReLU function to mitigate over-confident prediction on the known classes, and the COC loss can be formulated as: 
\begin{equation}
    \mathcal{L}_{coc}(x;\mathcal{P}) = \sum_C \max \left(\lambda_C - \theta, 0\right)
\end{equation}
By applying this penalty to the original classification loss, the model's overconfident predictions for unknown classes can be alleviated. Therefore, the COC loss can effectively mitigate the misclassification of unknown instances as known.

\subsection{Open-Set Prediction}
Finally, our optimization function is formulated as:
\begin{equation}
    \mathcal{L} = \mathcal{L}_{hc}(x ; \mathcal{P})+ \alpha \mathcal{L}_{amc}(x ; \mathcal{P}, \mathcal{R}) + \beta\mathcal{L}_{coc}(x;\mathcal{P})
\end{equation}
where we set $\alpha = \beta = 0.1$ empirically. The ablation study on the hyper-parameter setting can be found in Section~\ref{sec:ablation}.

We follow the open-set prediction strategy in~\cite{chen2021adversarial}. The probability of sample $x$ belonging to class $C$ is proportional to the distance between its feature embeddings $F(x)$ and the corresponding farthest reciprocal point of class $C$. For the closed-set classification tasks, we directly extract the class with the maximum distance. To assess the model's ability to detect unknown samples, we compute the difference between known and unknown probabilities for the test samples. This approach eliminates the need for careful threshold adjustment, providing a robust and unbiased assessment.

\section{EXPERIMENT}
\label{sec:exper}

\subsection{Datasets}
\label{sec:dataset}
\subsubsection{JIGSAWS Dataset}
We first evaluate our proposed framework on the publicly accessible  JIGSAWS dataset~\cite{gao2014jhu}.
JIGSAWS dataset includes kinematic and video data from eight surgeons executing five repetitions of three basic surgical tasks with the da Vinci surgical system. 
Three sub-datasets (suturing, knot tying, and needle passing) are standard components of most surgical skills training curricula. Fifteen surgical activity categories were defined and annotated. We use the official Leave-One-Supertrial-Out (LOSO) 5-fold cross-validation following~\cite{gao2014jhu}, and report the average results over the five folds. 
We repartition the JIGSAWS dataset to make it suitable for OSR tasks, and sample certain classes as unknown, as shown in Table~\ref{tab:jigsaws_distribution}. 

\subsubsection{DREAMS Dataset}

\begin{table}[h]
\renewcommand{\arraystretch}{1.5}
\caption{Class distribution of known and unknown classes in JIGSAWS dataset.}
 \centering
\label{tab:jigsaws_distribution}  
\resizebox{0.48\textwidth}{!}{	
\begin{tabular}{c|c|c|c}
\toprule[1pt]

Classes      & Needle Passing     & Knot Tying       & Suturing      \\ \hline
Known & G1, G3, G4, G5, G6   & G1, G11, G12, G15  & G1, G2, G3, G6, G8, G11        \\ \hline
Unknown & G2, G8, G11   & G13, G14  & G4, G5, G9, G10        \\  
\bottomrule[1pt]
\end{tabular}}
\end{table}

\begin{figure*}[t]
    \centering
    \includegraphics[width=0.9\linewidth, trim=10 160 0 0]{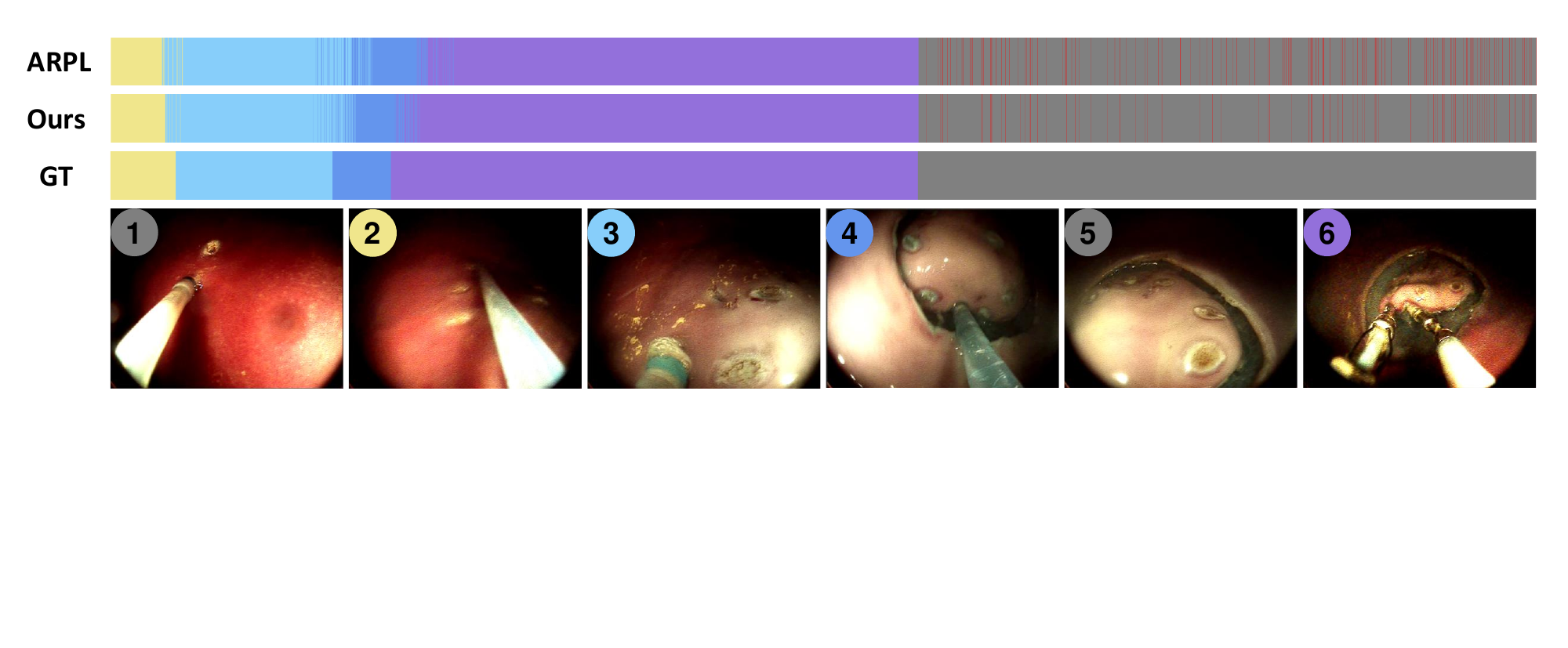}
    \caption{Color-coded ribbon illustration for DREAMS dataset. The gray band depicts the unknown classes, whereas the others denote the known classes. The red indicates the unknown samples that the model fails to recognize correctly. The six classes are marking, injection, circumferential incision, subsidized injection, installation and debugging, and bimanual submucosal dissection (with order).}
    \label{fig:vis}
\end{figure*}

\begin{table*}[t]
\caption{Comparison of experiment results against SOTA solutions on the JIGSAWS dataset.}
\centering
\label{tab:jigsaws}  
\resizebox{0.85\textwidth}{!}{
\begin{tabular}{c|ccc|ccc|ccc}
\toprule[1pt]

Task               & \multicolumn{3}{c|}{Needle Passing}              & \multicolumn{3}{c|}{Knot Tying}                  & \multicolumn{3}{c}{Suturing}                \\ \hline
Known/Unknown & \multicolumn{3}{c|}{5/3}                          & \multicolumn{3}{c|}{4/2}                          & \multicolumn{3}{c}{6/4}  \\ \hline
Metrics               & Acc            & AUROC          & OSCR           & Acc            & AUROC          & OSCR           & Acc            & AUROC          & OSCR     \\ \hline
OpenMax~\cite{bendale2016towards}               & 53.99          & 52.49          & 34.30          & 65.48          & 59.46          & 54.44          & 70.69          & 66.39          & 54.82 \\ 
RPL~\cite{chen2020learning}                 & 61.48          & 56.61          & 37.35          & 81.38          & 60.72          & 55.06          & 83.02          & 68.97          & 56.05 \\ 
DIAS~\cite{moon2022difficulty}               & 59.74          & 55.75          & 37.87          & 81.99          & 66.64          & 59.97          & 81.39          & 70.04          & 60.84  \\ 
DEAR~\cite{bao2021evidential}            & 62.04          & 56.80          & 38.16          & \textbf{82.33}          & 62.92          & 59.87          & 82.87          & 68.10          & 56.37       \\ 
ARPL~\cite{chen2021adversarial}              & 62.49          & 58.04          & 39.62          & 81.85          & 69.16          & 60.33          & 83.32          & 77.64          & 67.71    \\ 
ARPL+CS~\cite{chen2021adversarial}         & 53.69          & 53.86          & 34.08          & 66.30          & 59.76          & 53.63          & 71.61          & 63.42          & 55.36   \\
MLS~\cite{vaze2021open}               & 61.64          & 56.43          & 37.09          & 81.72          & 68.38          & 60.82          & 81.08          & 70.94          & 60.48   \\\hline 
Ours w/o $\mathcal{L}_{hc}$ & 61.80          & 59.44          & 40.44          & 81.67          & 68.69          & 60.84          & 82.84    & 78.84   & 68.87 \\ 
Ours w/o $\mathcal{L}_{coc}$ & 62.62          & 59.11          & 40.24          & 81.93          & 71.40          & 61.16          & 83.28    & 77.60   & 68.53 \\\hline 
Ours                  & \textbf{64.33} & \textbf{60.27} & \textbf{41.73} & 82.32 & \textbf{71.98} & \textbf{61.49} & \textbf{83.52} & \textbf{79.33} & \textbf{69.74} \\ \bottomrule[1pt] 
\end{tabular}}
\end{table*}

We present the DREAMS (Dual-arm Robotic Endoscopic Assistant for Minimally invasive Surgery) dataset, which is a challenging surgical dataset focusing on endoscopic submucosal dissection (ESD). The animal study was approved by the Institutional Ethics Committee on Animal Experiments (Approval No. DWLL-2021-021). Using our developed robotic system~\cite{gao2023transendoscopic,yang2023novel}, we collected 21 videos of complete robotic ESD on in-vivo porcine models, with 18 videos allocated to the training set and the remaining as the test. The ESD videos are recorded using a flexible dual-channel endoscope (Smart GS-60DQ, HUACO, China) at a frame rate of 30 FPS, with a resolution of $1920 \times 1080$. After cropping, the endoscopic image resolution is 
$ 1300 \times 1024 $. Following video preprocessing, expert endoscopists from Qilu Hospital provide detailed annotations of the ESD activities. The ESD activities encompass six classes:
marking, injection, circumferential incision, subsidized injection, installation and debugging, and bimanual submucosal dissection. For the purpose of OSR tasks, we designated class No.1 and No.5 as unknown, with the visualization in Fig.~\ref{fig:vis}.

\subsection{Implementation Details}
\label{sec:implementation}

We compare our proposed solution against the following SOTA solutions: OpenMax~\cite{bendale2016towards}, RPL~\cite{chen2020learning}, DIAS~\cite{moon2022difficulty}, DEAR~\cite{bao2021evidential}, ARPL~\cite{chen2021adversarial}, ARPL+CS~\cite{chen2021adversarial}, and MLS~\cite{vaze2021open}. All models are trained using Adam~\cite{kingma2014adam} for $90$ epochs with PyTorch on RTX 3090 GPU. We set the batch size to $64$ and the learning rate to $1 \times 10^{-5}$. 
We use closed-set accuracy, AUROC~\cite{neal2018open}, and OSCR score~\cite{dhamija2018reducing} as evaluation metrics. 

\subsection{Experimental Results}
\label{sec:results}
\subsubsection{Evaluation on JIGSAWS}

\begin{table}[t]
\caption{Comparison of experiment results against SOTA solutions on the DREAMS dataset.}
\centering
\label{tab:dreams}
\begin{tabular}{c|ccc}
\toprule[1pt]
Dataset                & \multicolumn{3}{c}{DREAMS} \\ \hline
Known/Unknown          & \multicolumn{3}{c}{4/2} \\ \hline
Metrics                & Acc    & AUROC  & OSCR  \\ \hline
OpenMax~\cite{bendale2016towards}                & 80.56  & 72.18  & 61.03 \\
RPL~\cite{chen2020learning}                    & 87.52  & 80.81  & 72.13 \\
DIAS~\cite{moon2022difficulty}                   & 89.59  & 86.65  & 81.87 \\
DEAR~\cite{bao2021evidential}                   & 88.34  & 79.75  & 69.82 \\
ARPL~\cite{chen2021adversarial}                   & 88.83  & 88.05  & 80.49 \\
ARPL+CS~\cite{chen2021adversarial}                & 81.10  & 77.89  & 67.85 \\
MLS~\cite{vaze2021open}                    & 89.06  & 86.24  & 80.56 \\ \hline
Ours w/o $\mathcal{L}_{hc}$             &   88.64 & 87.97       &   80.64    \\
Ours w/o $\mathcal{L}_{coc}$ & 89.95 & 87.63 & 80.91 \\ \hline
Ours                   & \textbf{90.88}  & \textbf{88.11}  & \textbf{82.21} \\
\bottomrule[1pt]
\end{tabular}
\end{table}

We present the quantitative experimental results on the JIGSAWS dataset in Table~\ref{tab:jigsaws}. Our methods demonstrate outstanding performance across all three surgical tasks, thereby showcasing substantial advancements in surgical activity recognition. Specifically, for the Needle Passing, Knot Tying, and Suturing tasks, our methods achieve accuracy of 64.33\%, 82.32\%, and 83.52\%, respectively. These results manifest a significant improvement compared to the ARPL, with percentage gains of 3.84\%, 2.82\%, and 1.81\%, respectively. This indicates the enhanced discriminative ability of our methods in distinguishing between known classes. Moreover, our methods demonstrate improved recognition capabilities for unknown classes, as evidenced by the highest attained AUROC score of 79.33\% in the Suturing task. Notably, the observed highest OSCR value across all tasks further highlights the exceptional proficiency of our model in precisely differentiating between known and unknown classes. These findings substantiate the effectiveness of leveraging COC loss and hyperspherical reciprocal points. By appropriately calibrating the over-confidence on the closed-set and facilitating a distinct separation between known and unknown classes, our OSSAR framework effectively cultivates a more discriminative feature space.

\subsubsection{Evaluation on DREAMS}
For the DREAMS dataset, we illustrate the quantitative results in Table~\ref{tab:dreams} and the qualitative results in Fig.~\ref{fig:vis}. Compared to SOTA OSR techniques, including the well-established ARPL~\cite{chen2021adversarial}, DIAS~\cite{moon2022difficulty}, and MLS~\cite{vaze2021open}, our approach consistently outperforms them across all three metrics.
These results showcase the efficacy of our approach in accurately classifying samples within the known classes while robustly detecting unknown instances. 

\subsection{Ablation Studies}
\label{sec:ablation}

\begin{table}[t]
\caption{Ablation study on the threshold $\theta$ of $\mathcal{L}_{coc}$.}
\centering
\label{tab:theta}
\begin{tabular}{c|ccc}
\toprule[1pt]
\multicolumn{1}{l|}{}       & \multicolumn{3}{c}{Needle Passing}          \\ \cline{2-4}
\multicolumn{1}{l|}{\multirow{-2}{*}{$\theta$}}   & Acc            & AUROC          & OSCR  \\  \hline
0                                     & 60.32 & 57.20 & 37.67 \\
5                                     & 61.69          & 59.04          & 38.85          \\
10                                    & \textbf{64.33} & \textbf{60.27} & \textbf{41.73} \\
15                                    & 62.20          & 59.75          & 40.57          \\
20                                    & 62.78          & 58.48          & 39.53 \\
\bottomrule[1pt]
\end{tabular}
\end{table}

\begin{table}[t]
\caption{Ablation on the weight parameters of the loss function.}
\centering
\label{tab5}
\begin{tabular}{cc|ccc}
\toprule[1pt]
\multicolumn{1}{c}{\multirow{2}{*}{$\alpha$}}  & \multicolumn{1}{c|}{\multirow{2}{*}{$\beta$}}  & \multicolumn{3}{c}{Needle Passing}                                \\ \cline{3-5}
 &  & Acc                  & AUROC                & OSCR                 \\ \hline
0.05 & 0.05 & 62.04 & 58.40 & 38.10 \\
0.05 & 0.1 & 61.11 & 58.72 & 39.21 \\
0.1 & 0.05  & 61.51 & 59.53 & 40.14 \\
0.1 & 0.1 & \textbf{64.33} & \textbf{60.27} & \textbf{41.73}       \\
0.1 & 0.5 & 61.49  & 57.69 & 38.85                \\
0.5 & 0.1 & 61.82 & 57.35 & 37.87\\
0.5 & 0.5 & 61.55 & 58.21 & 39.02 \\
\bottomrule[1pt]
\end{tabular}
\end{table}

\begin{table}[t]
\caption{Ablation study on the distance metrics for adversarial margin constraint.}
\centering
\label{tab:distance}
\begin{tabular}{c|ccc}
\toprule[1pt]
\multicolumn{1}{l|}{}       & \multicolumn{3}{c}{Needle Passing}          \\ \cline{2-4}
\multicolumn{1}{c|}{\multirow{-2}{*}{Distance Metric}}   & Acc            & AUROC          & OSCR  \\  \hline
\multicolumn{1}{c|}{Euclidean} & \textbf{64.33} & \textbf{60.27} & \textbf{41.73} \\
\multicolumn{1}{c|}{Angular}   & 62.57          & 55.57          & 38.48 \\
\multicolumn{1}{c|}{Manhattan}   & 57.10          & 60.25          & 37.56 \\
\multicolumn{1}{c|}{Chebyshev}   & 53.83          & 58.97          & 34.05 \\
\bottomrule[1pt]
\end{tabular}
\end{table}

\noindent \textbf{Effects of the effectiveness of proposed modules.} Table~\ref{tab:jigsaws} and~\ref{tab:dreams}  investigate the effectiveness of each proposed loss function in our solution. Specifically, we (i) degrade the $\mathcal{L}_{hc}$ to $\mathcal{L}_c$ and (ii) remove the $\mathcal{L}_{coc}$ to observe the performance. The quantitative results indicate that the exclusion of either component will adversely affect the overall performance. This observation evidences the positive contributions made by both proposed modules towards achieving the best results.

\noindent \textbf{Effects of the threshold $\theta$ of $\mathcal{L}_{coc}$.} In Table~\ref{tab:theta}, we select the best result, i.e. $\theta=10$, as our final model. When $\theta=0$, we can find that the model points in the non-informative direction. When other values are selected, the model's performance declines slightly, yet it still achieves comparable results to those of ARPL.

\noindent \textbf{Effects of loss weight parameters.} In Table~\ref{tab5}, we explore the effect of different loss weights $\alpha$ and $\beta$ on the model's performance by gridding, and end up with $\alpha = \beta = 0.1$ to get the best performance. 

\noindent \textbf{Effects of the distance metric of $\mathcal{L}_{amc}$.} In Table~\ref{tab:distance},
we further investigate the distance metric in $\mathcal{L}_{amc}$. 
When employing angular distance, AMC exhibits improved effectiveness compared to the use of Euclidean, Manhattan, or Chebyshev distances, which substantiates our proposed method in Section~\ref{sec:hsrp}.

\subsection{Discussion and Limitations}
In general, our proposed methodology has demonstrated significant advancements in performance across various comparative experiments. It effectively addresses the challenges of classifying known classes and identifying unknown classes. This accomplishment is achieved by segregating known and unknown classes within the feature space using hyperspherical reciprocal points. Additionally, the calibration process mitigates the problem of excessive confidence in classifying unknown categories. However, though adopting hyperspherical feature space introduces novel angular computations for distance measurement, we still require the utilization of Euclidean distance to set the margin constraint. This setting does not fully align with and leverage the characteristics of hyperspherical space.
Therefore, future research efforts will be directed towards dynamically combining angular and Euclidean distances to achieve exceptional adaptability in OSSAR and other open-set surgical tasks.

Furthermore, our approach does not exhibit substantial enhancements in closed-set accuracy; instead, it primarily concentrates on improving open-set task performance. Considering the high-risk nature of surgical robotics, the current level of accuracy falls short of meeting the demands of next-generation automated robotic surgery, which serves as the focal point of our future endeavors.

\section{CONCLUSIONS}
\label{sec:conclusion}
This paper introduces the OSSAR framework to solve the open-world challenges in surgical activity recognition. Specifically, the HSRP strategy is proposed to improve the clustering and separation between features of known and unknown classes. The COC loss is employed to mitigate the misclassification of unknown classes as known classes due to excessive model confidence in known classes. Subsequently, we establish the OSSAR benchmark on two robotic surgery datasets, JIGSAWS and DREAMS. Extensive experiments demonstrate the effectiveness of our proposed method and each individual component. Our approach achieves outstanding performance in both closed-set accuracy and unknown class detection. Future works shall include striving for further improvements in closed-set performance and exploring optimal combinations of linear and angular distance. Our efforts will also be dedicated to deploying our algorithm in surgical robot systems to advance surgical robot automation.







\balance
\bibliographystyle{IEEEtran}
\bibliography{reference}

\end{document}